\documentclass[12pt]{iopart}
\pdfoutput=1
\usepackage{graphicx}

\usepackage{dcolumn}
\usepackage{amssymb}
\usepackage[T1]{fontenc}
\usepackage[applemac]{inputenc}
\usepackage{subfigure}
\usepackage{geometry}
\usepackage{array}
\usepackage{color}
\usepackage{algpseudocode}

\begin{document}

\title{A novel local search based on variable-focusing for random K-SAT}

\author{R\'{e}mi Lemoy$^{1,2}$, Mikko Alava$^{2}$, Erik Aurell$^{2,3}$}

\address{$^{1}$Aalto University, Department of Applied Physics, Espoo, Finland,\\
$^{2}$Aalto University, Department of Information and Computer Science, Espoo, Finland,\\
 $^{3}$Department of Computational Biology, and ACCESS Linnaeus Centre, KTH-Royal Institute of Technology, Stockholm, Sweden}

\eads{\mailto{remilemoy@gmail.com}, \mailto{mikko.alava@aalto.fi}, \mailto{eaurell@kth.se}}

\date{\today}

\begin{abstract}
We introduce a new local search algorithm for satisfiability problems.
Usual approaches focus uniformly on unsatisfied clauses. The new method works by picking uniformly random variables in unsatisfied clauses.
A Variable-based Focused Metropolis Search (V-FMS) is then applied to random 3-SAT. We show that it is quite comparable in performance to
the clause-based FMS. Consequences for algorithmic design are discussed.
\end{abstract}



\section{Introduction}
Focused local search algorithms have been found to be fairly efficient in solving random instances of the Boolean satisfiability problem (K-SAT), a famous NP-complete problem for $K \geq 3$ \cite{Garey1979}. The first, simplest but still very interesting,
such algorithm is the one introduced by Papadimitriou in 1991 \cite{Papadimitriou1991} where variables are flipped
randomly, independently and at a rate proportional to how many unsatisfied clauses they each participate in.
Randomness and greediness in focused local search were combined in the Walksat algorithm of
Selman, Kautz and Cohen~\cite{SKC}, and several other variants in this direction have been
developed and investigated~\cite{Schoning1999,Gordon2005,seitz05,Ardelius2006,alava07,Kroc2010}.
A physics-based outlook on the question of finding solutions follows from considering K-satisfiability as a diluted 3-spin Ising model with disorder \cite{Martin2001,Montanari2008}. The variables of SAT formulas are then considered as spins, and the clauses, which consist each of $K$ literals (i.e. variables or their negations), are local energy terms. This glassy model has been analyzed by usual techniques and its phase diagram worked out. However, focused local search does not obey detailed
balance since all zero-energy states are left unchanged by this family of algorithms, and equilibrium considerations
may therefore not be very relevant for the behaviour or design of such algorithms. Indeed, it is well known that the
equilibrium phase diagram of random K-SAT does not determine when focused local search works or does not work, and
such information, although very important in itself, has not given much of a hint on how to proceed with the development of efficient local algorithms.

In this, Focused Metropolis Search (FMS) and its variants \cite{seitz05,alava07,Ardelius2006} have been a main empirical step forward.
These algorithms display a linear scaling of the solution times with instance size even in the immediate proximity of the SAT/UNSAT transition \cite{seitz05,alava07}, identified by the ratio $\alpha=M/N$ of the number of clauses $M$ to the number of variables $N$ \cite{Mitchell1992}. Similarly to the Walksat, focusing on unsatisfied clauses during the search is an important ingredient of this succes. It is implemented as in~\cite{Papadimitriou1991}, where first a random choice is made among unsatisfied clauses, and then among the variables participating in the chosen clause.

Thus in the FMS the rate of picking a variable is not uniform among the set of variables participating in unsatisfied clauses, but is biased towards variables participating in many unsatisfied clauses. In this work,  we introduce the symmetric approach of uniform focusing on variables in unsatisfied clauses.
This is a natural choice since it means a uniform measure on the subset of spins that contribute to the energy. We investigate the Variable-based Focused Metropolis Search, i.e. V-FMS, with the idea of showing that upon an intuitive rescaling of solution times it performs at least as well as the usual FMS on 3-SAT. The fact that the kind of sampling is irrelevant and a number of other observations should be of importance for choosing further directions in statistical physics -based algorithms: their development and understanding both empirically and theoretically.

The structure of the rest of this work is as follows.
Below, we present in pseudo-language the Variable-based Focused Metropolis Search, i.e. V-FMS as an example of variable-based focusing. Then, we demonstrate in Section 3 that the algorithm can be compared directly to the FMS results utilizing earlier data for FMS published in Ref.~\cite{seitz05}. The section also contains several new observations about algorithmic behavior. Section 4 finishes the paper with conclusions.

\section{Variable-based focusing}
In what follows, the state of the system at a given time is monitored by the number of unsatisfied clauses in the configuration (i.e. assignment of values to the variables), which we use as an energy function. It reaches zero only when a solution is found, and can be used to define an energy landscape~\cite{Krzakala2007} (but see also below). We study the variable-focusing case with the behavior of the Focused Metropolis Search (V-FMS) algorithm, presented here in pseudocode:

\begin{algorithmic}[1]
\State $S=$ random assignment of values to the variables
\While {$S$ is not a solution}
\State $V=$ a variable selected uniformly from those in UNSAT clauses
\State $\Delta E=$ change in energy if $V$ is flipped in $S$
\If {$\Delta E \leq 0$}
\State flip $V$ in $S$
\Else \State flip $V$ in $S$ with probability $\eta^{\Delta E}$
\EndIf
\EndWhile
\end{algorithmic}

We have implemented the above code as new search heuristics in a code derived from Walksat version 45~\cite{Walksat,SKC,alava07},
where we have introduced a new array to keep track of variables participating in unsatisfied clauses.
In this code, $\eta$ is a positive parameter which has to be adjusted for optimal search.
Let us point out that since this process does not obey detailed balance the ``energy'' $E$ is here just a convenient quantity to describe the dynamics locally, but which does not necessarily have anything
to say about the dynamics globally. Instead of $E$, and given that V-FMS concentrates on variables, it is an alternative and natural idea to consider
$N_u$, the number of variables involved in at least one unsatisfied clause (the cardinality of the support set of the flipping).
This quantity also defines an energy landscape which is zero only at solutions, which could in principle have qualitatively
different properties, and one could use this or other quantities to define the best energy
landscapes for the task one is interested in. 

\section{Algorithmic performance}
\subsection{Role of noise for V-FMS}
Next we test the V-FMS against the known state-of-the-art of the FMS, and we refer in particular to the work of Seitz et al. \cite{seitz05}. The important issues here from an empirical perspective are: i) is the V-FMS linear for relatively high constraint-densities $\alpha$? ii) how does the performance depend on the noise parameter $\eta$, i.e. what does the algorithmic landscape look like? Ordinary FMS is biased towards picking trial variables that are present in more than average number of unsatisfied clauses. If this bias and the evolution of the set of eligible variables does not matter, then one should get similar performance out of V-FMS.

\begin{figure}[h!]
\centering
\begin{tabular}{cc}
\includegraphics[width=8cm]{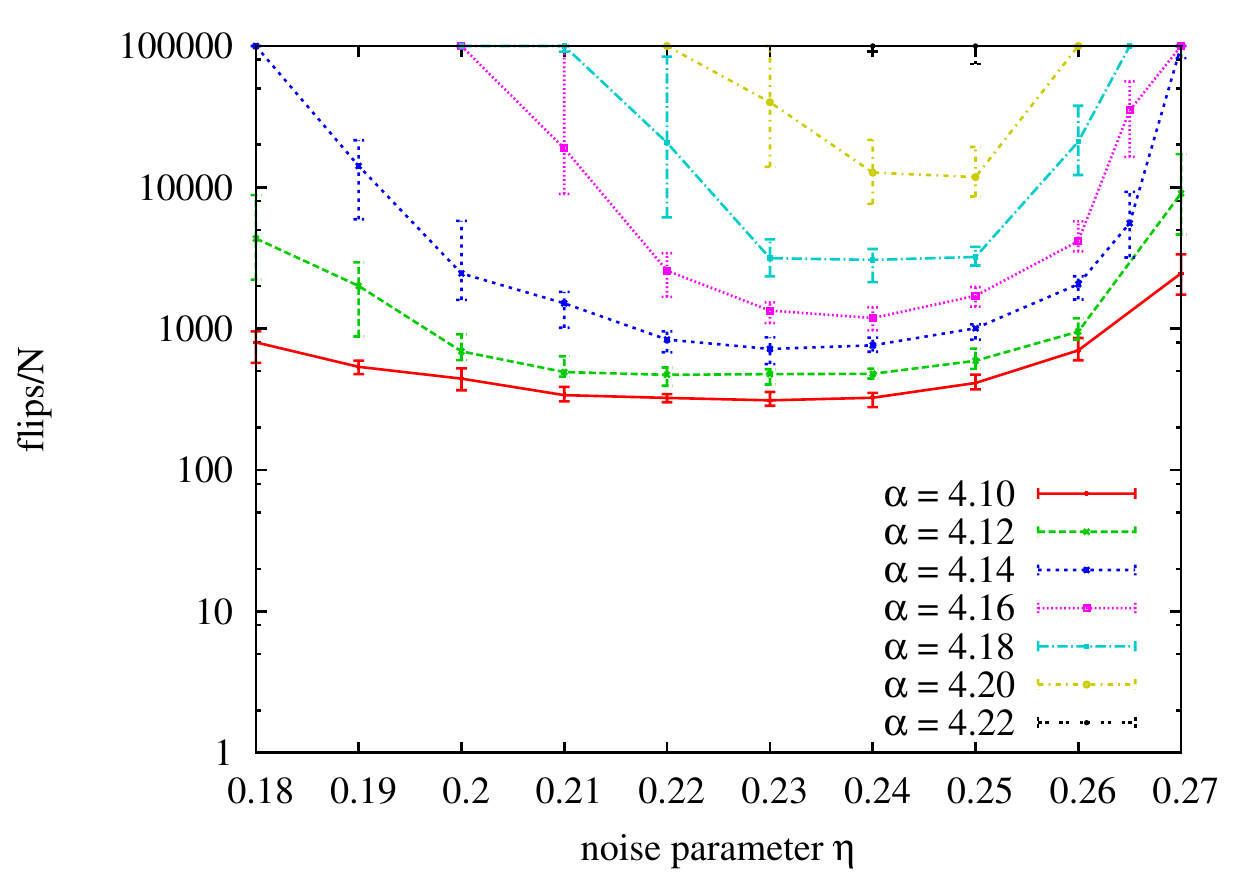} &
\includegraphics[width=8cm]{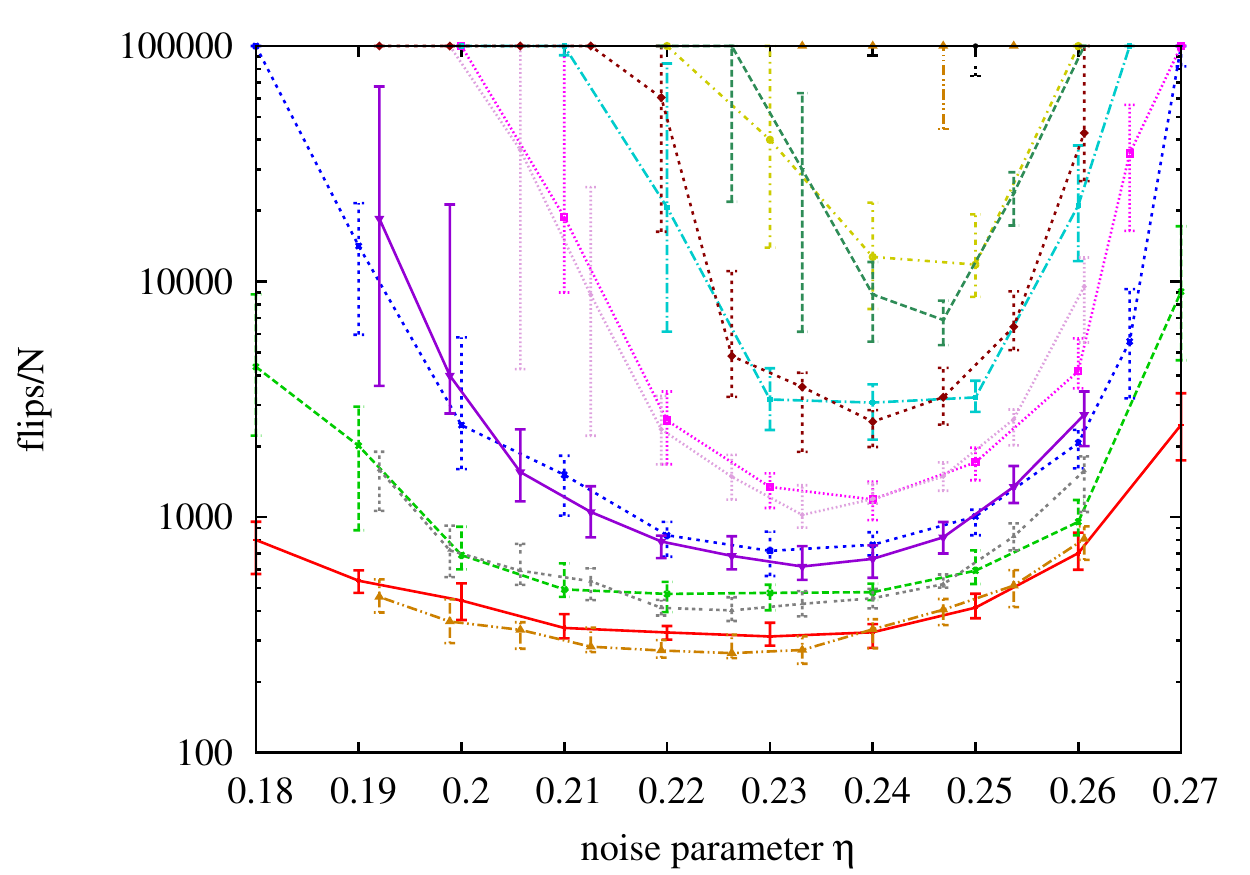}
\end{tabular}
\caption[Running times]{Running times of the local search algorithms. On the left panel, running time of the V-FMS algorithm as a function of the noise parameter $\eta$, for different values of $\alpha$. Each point presents median and quartiles for 20 instances, with $N=10^5$. The right panel presents these results again, with the same symbols, and superimposes the running times of the FMS algorithm taken from \cite{seitz05} for the same values of $\alpha$, but where the noise parameter $\eta$ has been rescaled by a factor $0.24/0.35$.}
\label{runtimes_feta}
\end{figure}

Indeed, despite the different sampling rules used for V-FMS and FMS, figure \ref{runtimes_feta} shows that the solution times are quite similar.
The generic feature is as is known for local search with focusing: the effort to solve problems in the median sense depends on the noise parameter $\eta$, or other such parameters of the algorithm used. There is a minimum solution time for an optimal noise parameter, and the minimum gets more marked with the increase of $\alpha$ as has been reported for the FMS in the past \cite{seitz05}. The right panel of figure \ref{runtimes_feta} shows that a simple rescaling of the noise parameter (data from Ref.~\cite{seitz05}) suffices to bring the curves of the solution times to fall well on top of each other. This implies that the particular sampling, clause or variable based, is at most of quantitative importance in these solving very difficult 3-SAT problems -- a more clear difference might well of course exist for other test cases, like $K>3$ in the SAT class. Let us note that multiplying the noise parameter $\eta$ by a constant is equivalent to multiplying the energy $E$ by a related constant. We can also observe that the optimal noise is higher in the case of the standard FMS, since that chooses preferably variables in several unsatisfied clauses, than in the case of VFMS, where variables in unsatisfied clauses are picked uniformly.

\begin{figure}[h!]
\centering
\includegraphics[width=8cm]{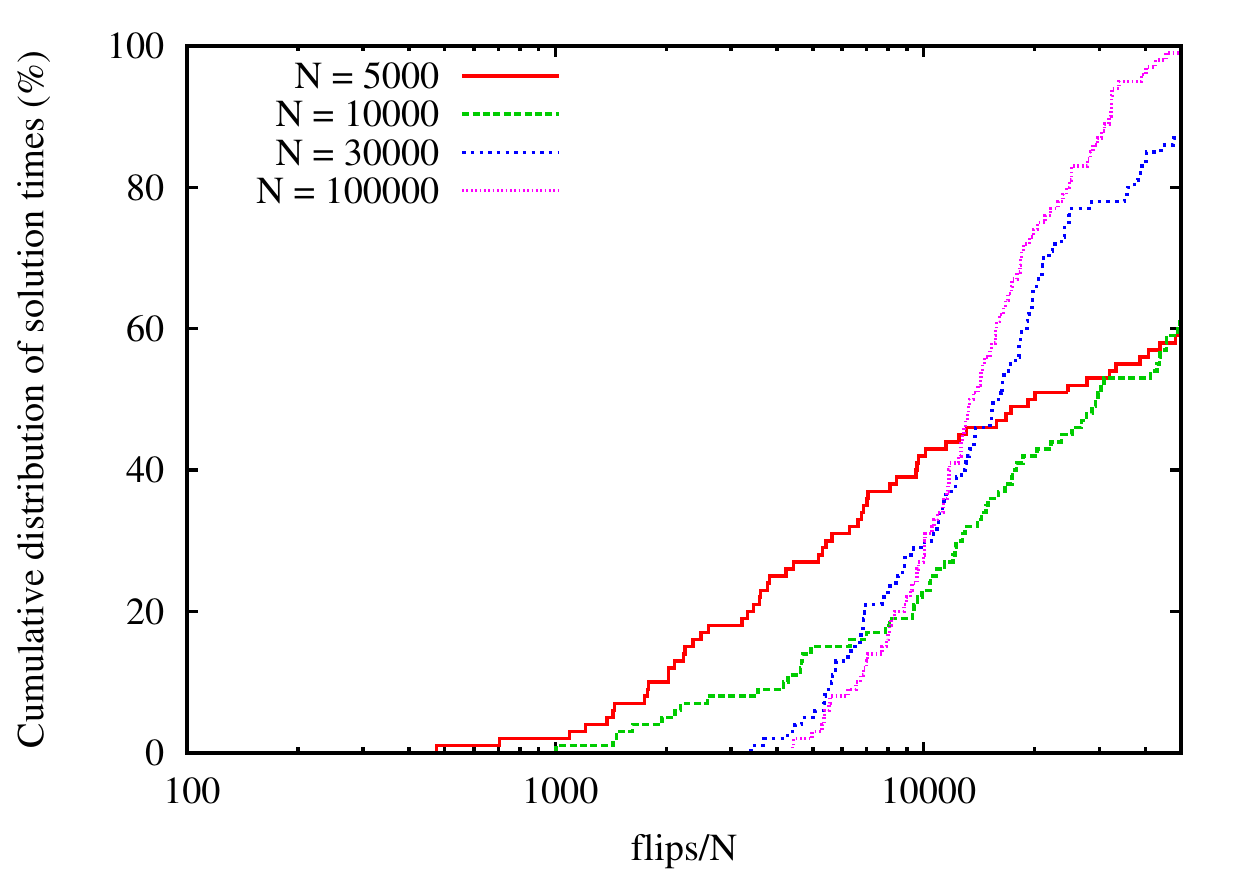}
\caption[Running times]{Cumulative distribution of running times of VFMS with $\alpha=4.2$ and $\eta=0.25$. Each curve corresponds to 100 instances.}
\label{runtimes}
\end{figure}

Figure \ref{runtimes}  affirms the natural expectation, that the solution time distribution gets more focused around the typical value with an increasing instance size. This concentration of the measure takes place for an $\alpha$ close to the maximum value for which the local search methods work (linear algorithms have been reported to perform at least up to $\alpha=4.23$ while Survey Propagation finds solutions for $\alpha=4.25$).

\subsection{Energy traces}
 Next we consider the choice of the energy landscape by investigating the combination of $N_u$ and the energy $E$. In the next figure \ref{traces}, we pick a suitable set of parameters including roughly optimal noise values for both V-FMS and FMS. We observe that both algorithms have close behaviors, but still distinct energy and $N_u$ traces. This can be related to Figure \ref{runtimes_feta}, where no difference was seen between algorithms on running times after the appropriate rescaling of the noise parameter $\eta$ is made. The data is presented both in linear and logarithmic timescales to underline the early time behaviors.
\begin{figure}[h!]
\centering
\begin{tabular}{cc}
\includegraphics[width=8cm]{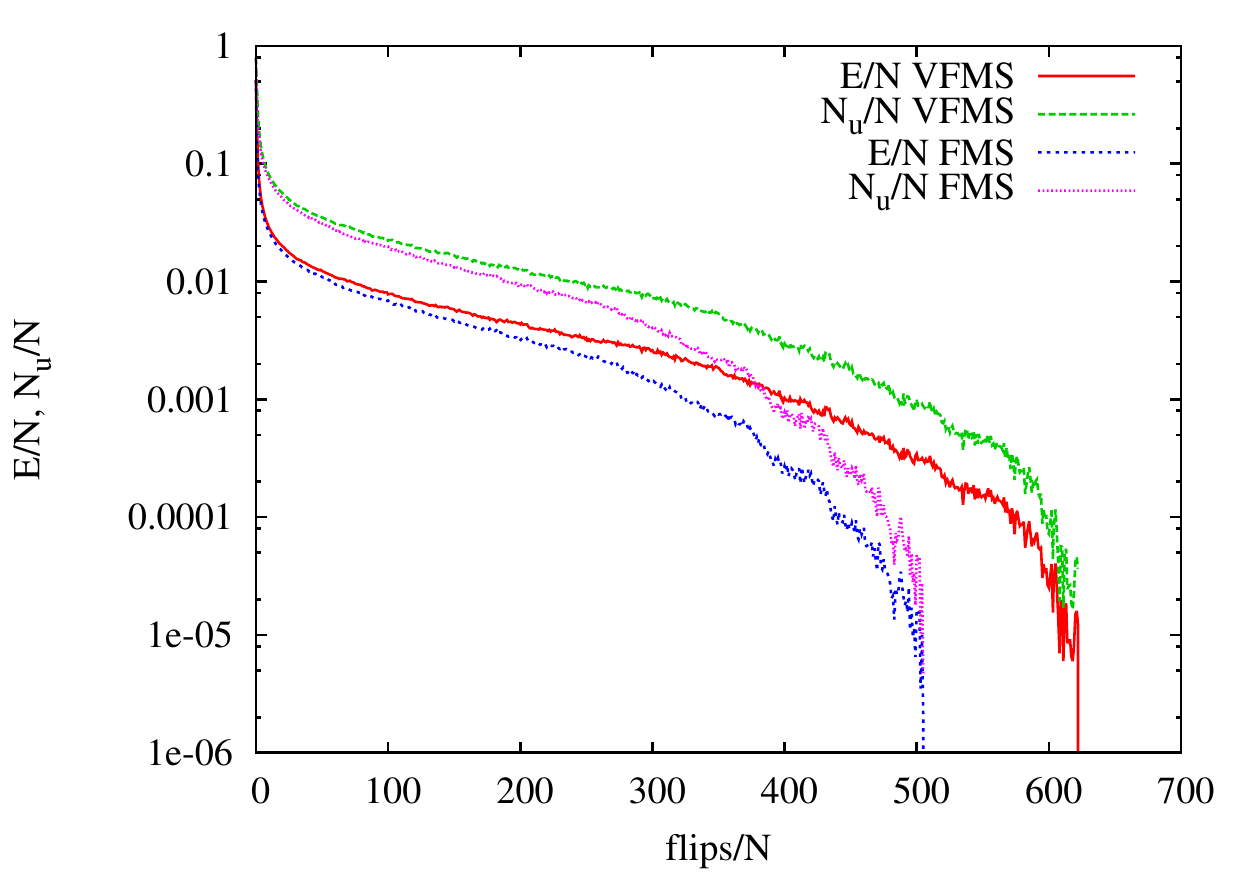} &
\includegraphics[width=8cm]{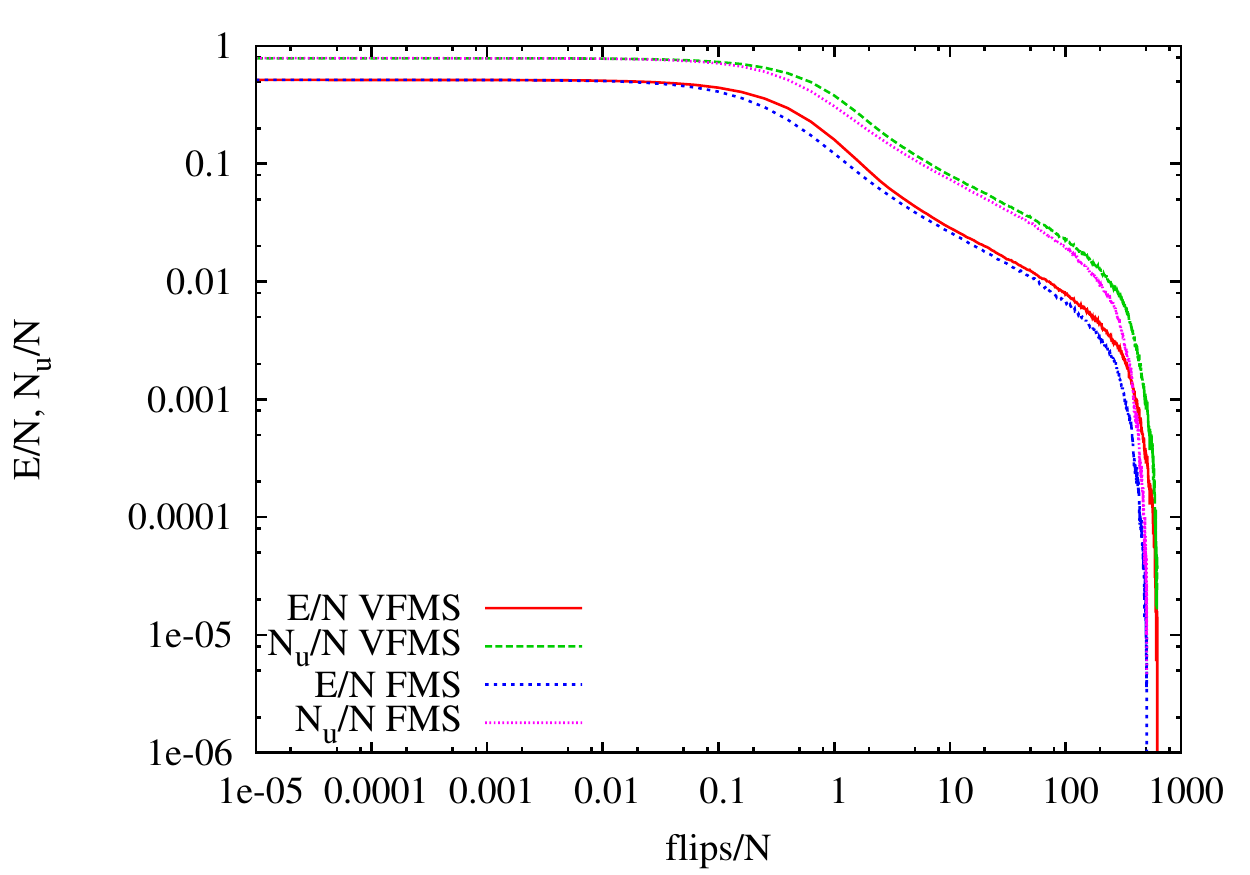} \\
\includegraphics[width=7.5cm]{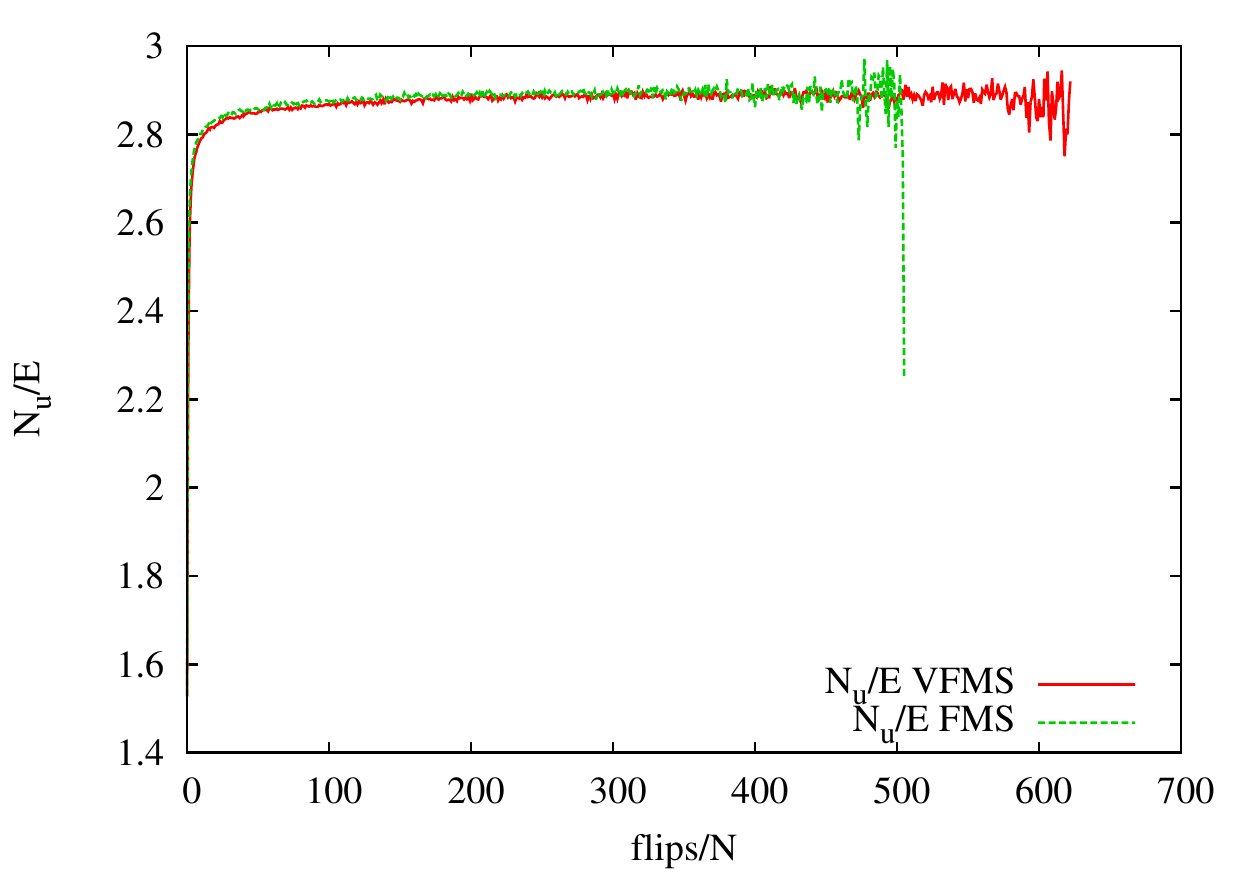} &
\includegraphics[width=7.5cm]{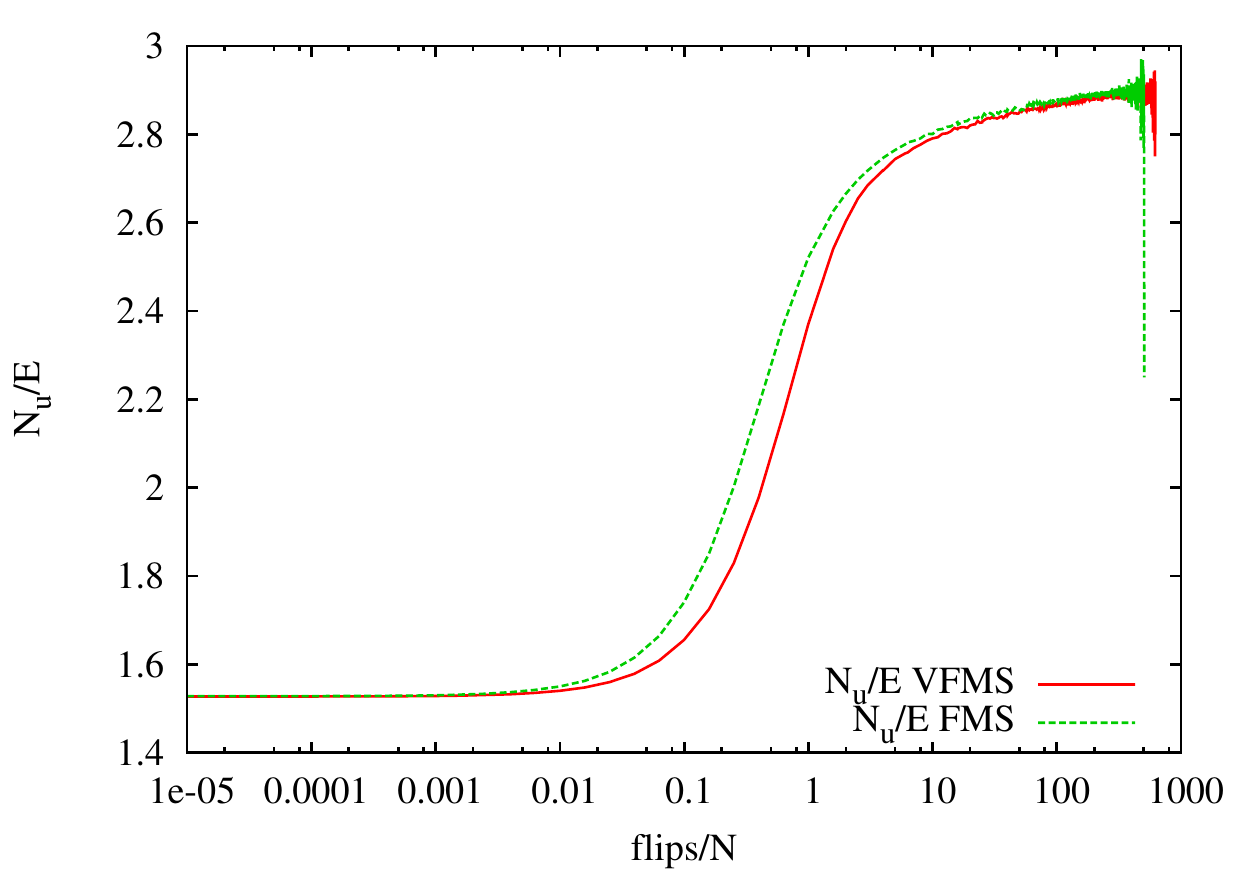} \\
\end{tabular}
\caption[Energy traces]{Top panel: average traces, over 20 instances, of the energy $E$ and the number of variables in unsatisfied clauses $N_u$, for FMS ($\eta=0.33$) and VFMS ($\eta=0.23$), with $\alpha=4.12$ and $N=10^5$. Bottom panel: same results, but showing the evolution of $N_u/E$.}
\label{traces}
\end{figure}

The main observation is that energy and number $N_u$ of variables in unsatisfied clauses have a very similar evolution. They are almost proportional during most of the search process evolution, as shown on the bottom panels of Figure \ref{traces}. The proportionality factor increases first and then saturates to a value slightly below $K=3$,  meaning that most variables involved in unsatisfied clauses are involved in only one of them. This of course helps to qualitatively understand why the different sampling does not bring about any major changes in performance (V-FMS compared to FMS). Note that for both algorithms there is a similar transient from the $N_u /E$ value of a random assignment towards the higher value; already roughly one flip per clause is enough for the transient to get over.

\section{Conclusion}
In this work, we have proposed the concept of focused search algorithms that concentrate on variable choice instead of constraint-based selection. This opens up the possibility 
of considering yet other versions of focusing such as \textit{e.g.}
to concentrate even more strongly on variables in many unsatisfied clauses, or to concentrate on variables
which participate in as few unsatisfied clauses as possible (but still at least one). It is an interesting question that should be pursued to determine if some of
these schemes can be implemented even more efficiently, as has been done here with V-FMS or what holds for FMS itself. The idea of variable-focusing arises naturally in
theoretical approaches based on describing the algorithm dynamics by a master equation, since the natural time scale for asynchronous updates is one over the number of variables \cite{mozeika-lemoy}.

One point we have raised is that many other quantities than the conventional ``energy'' $E$ can
be considered to design efficient local search, and we have pointed out that one natural
such quantity could be $N_u$, the number of variables in unsatisfied clauses. Some such
alternative quantities have already been considered in the past such as
``breakcount'' (number of clauses which become unsatisfied if a given variable is flipped) and ``makecount'' (number of clauses which become satisfied) in Walksat~\cite{SKC}, and the remembered value of
a previous energy minimum in focused record-to-record travel~\cite{seitz05};
$N_u$ has however the conceptual advantage that it is simpler
and similarly to $E$ can be naturally extended to a global landscape.
The tests for 3-SAT reveal that the V-FMS is in practice not better nor worse than the usual focused Metropolis search. Tests also revealed that $E$ and $N_u$ are 
proportional (or nearly so) after an initial transient, and a value is reached which indeed indicates that eligible variables tend to partake only in one unsatisfied clause. 
Since this is close to the "maximum efficiency"  for the ratio of the two quantities, it is an interesting question how to find algorithms that differ in this sense,
and if they would be found to be improved over (V)-FMS.

The apparent boundary of (typical) linear behaviour of focused local search on random K-SAT is hence approximately at the same value of $\alpha\approx 4.23$ with quite different choices of
local search, as has been found in the past~\cite{seitz05}, and also for different kinds of focusing, as
found here. Why this is, and if this boundary is in some sense universal,
or if other versions of focusing can be designed which perform qualitatively differently
are important challenges for the future.

\subsection*{Acknowledgements}
The authors thank Alexander Mozeika for discussions about variable focusing. Prof. Pekka Orponen (Aalto)
 is acknowledged for discussions. This work was supported by funding from the Center of Excellence program of the Academy of Finland, for the COIN and COMP Centers, 
 as well as the Academy of Finland Distinguished Professor program (E.A.).

\section*{References}
\bibliographystyle{unsrt}
\bibliography{biblio}

\begin{thebibliography}{10}

\bibitem{Garey1979}
Michael~R Garey and David~S Johnson.
\newblock {\em Computers and intractability}, volume 174.
\newblock Freeman New York, 1979.

\bibitem{Papadimitriou1991}
Christos~H. Papadimitriou.
\newblock On selecting a satisfying truth assignment.
\newblock In {\em Foundations of Computer Science, 1991. Proceedings., 32nd
  Annual Symposium on}, pages 163--169, 1991.

\bibitem{SKC}
Bart Selman, Henry Kautz, and Bram Cohen.
\newblock Local search strategies for satisfiability testing.
\newblock In David~S. Johnson and Michael~A. Trick, editors, {\em Cliques,
  Coloring, and Satisfiability: Second DIMACS Implementation Challenge},
  volume~26, pages 521--532. AMS, October 1996.

\bibitem{Schoning1999}
U.~Sch\"oning.
\newblock A probabilistic algorithm for k-sat and constraint satisfaction
  problems.
\newblock In {\em Proceedings of 40th Annual Symposium on Foundations of
  Computer Science}, pages 410--414, 1999.

\bibitem{Gordon2005}
Erik Aurell, Uri Gordon, and Scott Kirkpatrick.
\newblock Comparing beliefs, surveys and random walks.
\newblock In {\em Advances in Neural Information Processing Systems 17:
  Proceedings of the 2004 Conference}, volume~17, page~49. The MIT Press, 2005.

\bibitem{seitz05}
Sakari Seitz, Mikko Alava, and Pekka Orponen.
\newblock Focused local search for random 3-satisfiability.
\newblock {\em Journal of Statistical Mechanics: Theory and Experiment},
  2005(06):P06006, 2005.

\bibitem{Ardelius2006}
John Ardelius and Erik Aurell.
\newblock Behavior of heuristics on large and hard satisfiability problems.
\newblock {\em Phys. Rev. E}, 74:037702, 2006.

\bibitem{alava07}
Mikko Alava, John Ardelius, Erik Aurell, Petteri Kaski, Supriya Krishnamurthy,
  Pekka Orponen, and Sakari Seitz.
\newblock Circumspect descent prevails in solving random constraint
  satisfaction problems.
\newblock {\em Proceedings of the National Academy of Sciences},
  105(40):15253--15257, 2008.

\bibitem{Kroc2010}
Lukas Kroc, Ashish Sabharwal, and Bart Selman.
\newblock An empirical study of optimal noise and runtime distributions in
  local search.
\newblock In Ofer Strichman and Stefan Szeider, editors, {\em Theory and
  Applications of Satisfiability Testing -- SAT 2010}, volume 6175 of {\em
  Lecture Notes in Computer Science}, pages 346--351. Springer Berlin
  Heidelberg, 2010.

\bibitem{Martin2001}
Olivier~C Martin, R{\'e}mi Monasson, and Riccardo Zecchina.
\newblock Statistical mechanics methods and phase transitions in optimization
  problems.
\newblock {\em Theoretical computer science}, 265(1):3--67, 2001.

\bibitem{Montanari2008}
Andrea Montanari, Federico Ricci-Tersenghi, and Guilhem Semerjian.
\newblock Clusters of solutions and replica symmetry breaking in random k
  -satisfiability.
\newblock {\em Journal of Statistical Mechanics: Theory and Experiment},
  2008(04):P04004, 2008.

\bibitem{Mitchell1992}
David Mitchell, Bart Selman, and Hector Levesque.
\newblock Hard and easy distributions of sat problems.
\newblock In {\em Proceedings of the tenth national conference on Artificial
  intelligence}, pages 459--465. AAAI Press, 1992.

\bibitem{Krzakala2007}
Florent Krzakala and Jorge Kurchan.
\newblock Landscape analysis of constraint satisfaction problems.
\newblock {\em Phys. Rev. E}, 76:021122, Aug 2007.

\bibitem{Walksat}
Bart Selman and Henry Kautz.
\newblock Walksat home page, http://www.cs.rochester.edu/u/kautz/walksat/.

\bibitem{mozeika-lemoy}
A.~Mozeika, R.~Lemoy, et~al., in preparation.

\end{thebibliography}


\end{document}